\pdfoutput=1

\documentclass[11pt]{article}

\usepackage[preprint]{acl}

\usepackage{times}
\usepackage{latexsym}
\usepackage{amsmath}
\usepackage{amssymb}
\usepackage{amsfonts}
\usepackage[T1]{fontenc}

\usepackage[utf8]{inputenc}

\usepackage{microtype}

\usepackage{inconsolata}

\usepackage{graphicx}

\title{Unlocking the Future: Exploring Look-Ahead Planning Mechanistic Interpretability in Large Language Models}
\author{
 \textbf{Tianyi Men\textsuperscript{1,2}},
 \textbf{Pengfei Cao\textsuperscript{1,2}},
 \textbf{Zhuoran Jin\textsuperscript{1,2}},
 \textbf{Yubo Chen\textsuperscript{1,2}},
 \textbf{Kang Liu\textsuperscript{1,2}},
 \textbf{Jun Zhao\textsuperscript{1,2}}
\\
 \textsuperscript{1}The Laboratory of Cognition and Decision Intelligence for Complex Systems,\\
 Institute of Automation, Chinese Academy of Sciences, Beijing, China\\
 \textsuperscript{2}School of Artificial Intelligence, University of Chinese Academy of Sciences, Beijing, China
\\
 \small{
   \{tianyi.men, pengfei.cao, zhuoran.jin, yubo.chen, kliu, jzhao\}@nlpr.ia.ac.cn
 }
}

\begin{document}
\maketitle
\begin{abstract}
Planning, as the core module of agents, is crucial in various fields such as embodied agents, web navigation, and tool using. With the development of large language models (LLMs), some researchers treat large language models as intelligent agents to stimulate and evaluate their planning capabilities. However, the planning mechanism is still unclear. In this work, we focus on exploring the look-ahead planning mechanism in large language models from the perspectives of information flow and internal representations. First, we study how planning is done internally by analyzing the multi-layer perception (MLP) and multi-head self-attention (MHSA) components at the last token. We find that the output of MHSA in the middle layers at the last token can directly decode the decision to some extent. Based on this discovery, we further trace the source of MHSA by information flow, and we reveal that MHSA mainly extracts information from spans of the goal states and recent steps. According to information flow, we continue to study what information is encoded within it. Specifically, we explore whether future decisions have been encoded in advance in the representation of flow. We demonstrate that the middle and upper layers encode a few short-term future decisions to some extent when planning is successful. Overall, our research analyzes the look-ahead planning mechanisms of LLMs, facilitating future research on LLMs performing planning tasks.

\end{abstract}

\section{Introduction}

\begin{figure}[t]
  \includegraphics[width=\columnwidth, height=8cm]{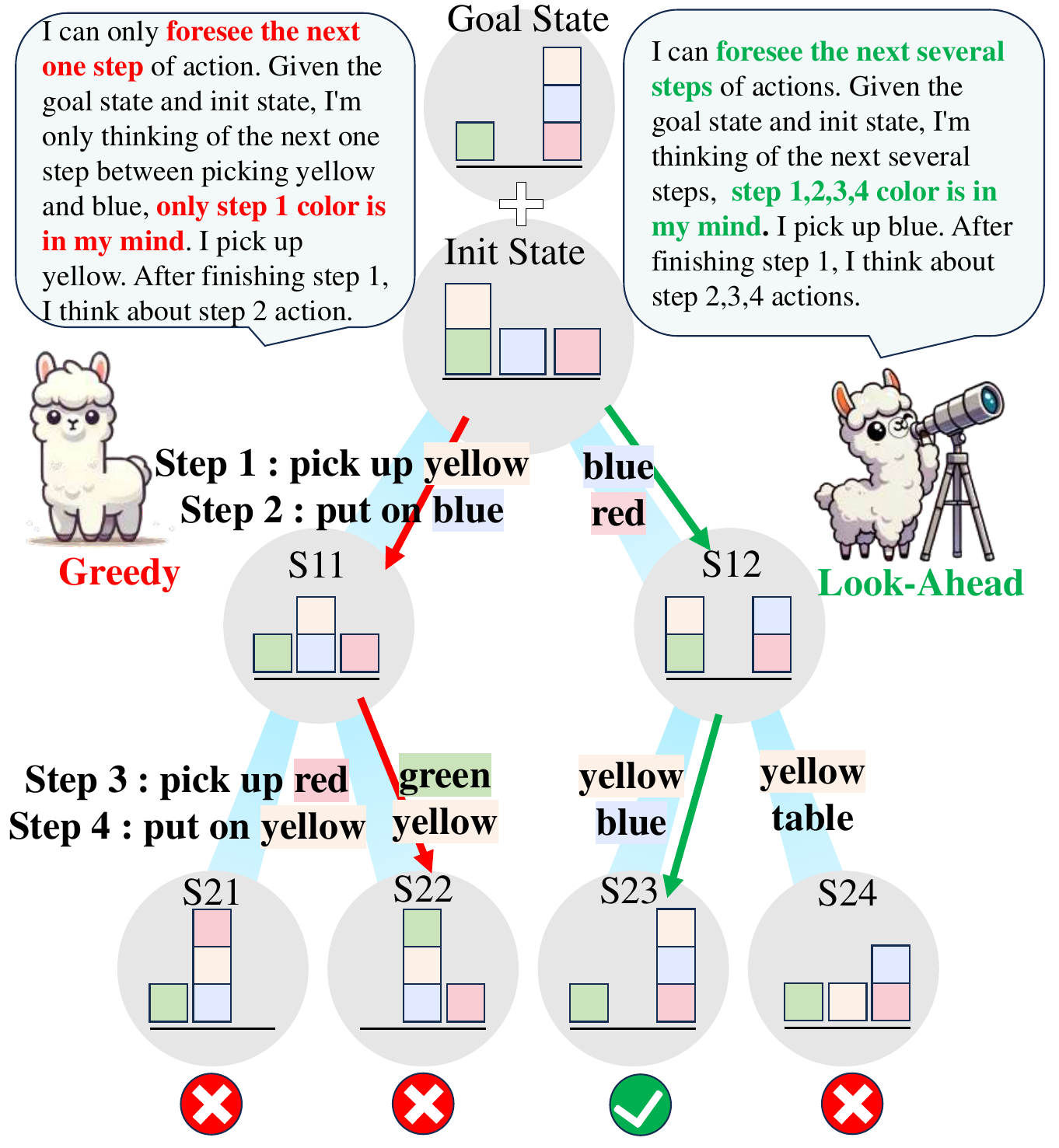}
  \caption{An example of greedy and look-ahead planning.}
  \label{fig_introduction}
\end{figure}

Planning is the process of formulating a series of actions to transform a given initial state into a desired goal state \cite{valmeekam2024planbench,zhang2024pddlego}. As the core module of agents \cite{xi2023rise,wang2024survey}, planning has been widely applied in many fields such as embodied agents \cite{shridhar2020alfworld,wang2022scienceworld}, web navigation \cite{zhou2023webarena,deng2024mind2web} and tool using \cite{xu2023tool,qin2023toolllm}. With the development of large language models, some researchers treat large language models as intelligent agents to solve complex tasks. This is because large language models may possess some preliminary planning capabilities \cite{huang2022language}. Recently, researchers have made efforts to stimulate and evaluate the planning capabilities of large language models. They propose prompt engineering \cite{zheng2023synapse} and instruction fine-tuning \cite{zeng2023agenttuning} to boost the planning abilities of large language models. Additionally, some researchers construct benchmarks such as AgentBench \cite{liu2023agentbench} and AgentGym \cite{xi2024agentgym} to evaluate the planning capabilities of large models. Although they have made some progress, the underlying mechanisms in planning capabilities of large language models remain a largely unexplored frontier. Revealing the planning mechanisms of large language models helps to better understand and improve their planning capabilities. Therefore, we focus on exploring the underlying mechanisms behind the planning abilities of large language models.

In this work, we focus on exploring look-ahead planning mechanisms in large language models. We study the classical planning task Blocksworld, which is a fully-observed setting. All entity states are known from the init state and goal state, so exploration is not needed \cite{zhang2024pddlego}. As illustrated in Figure~\ref{fig_introduction}, given an initial state and a goal state of Blocksworld, the model can only pick up or put down one block. The model must generate a sequence of actions to transform the initial state into the goal state, as shown by the green path. However, it is still unclear whether the model, at step $t$,  greedily considers only the action at $t+1$ or look-ahead considers the actions at $t+2$ and beyond. Inspired by psychology, humans engage in look-ahead thinking when making plans \cite{baumeister2016pragmatic}. Based on this, we further propose the hypothesis of model look-ahead planning, which is as follows:

\begin{itemize}
\item \textbf{Look-Ahead Planning Decisions Existence Hypothesis}: In the task of planning with large language models, given a rule, an initial state, a goal state, and task description prompts. At the current step, the model needs to predict the next action, the probe can detect decisions to some extent for future steps in the internal representations in the short term within a fully-observed setting when planning is successful.
\end{itemize}

We design a two-stage paradigm to verify this hypothesis. It can be divided into the finding information flow stage and the probing internal representations stage. The first stage is to analyze the information flow and component functions during planning (§\hyperref[sec:5]{5}). The second stage is examining whether the model stores future information in internal representations (§\hyperref[sec:6]{6}). The specifics are as follows:

(1) In the first stage, we study how planning is done internally by analyzing the MLP and MHSA components at the last token. Inspired by methods of calculating extraction rates methods \cite{geva2023dissecting}, we find the output of MHSA in the middle layers at the last token can directly decode the correct colors to some extent (§\hyperref[sec:51]{5.1}). Based on this discovery, we further investigate the sources of information on MHSA. We trace the source of the decisions. And find that planning mainly depends on spans of the goal states and recent steps (§\hyperref[sec:52]{5.2}). 

(2) In the second stage, we study what information is encoded in the information flow and whether this information has been considered in advance for future decisions. For future decisions existence, we use the probing method to probe future decisions and reveal that the middle and upper layers encode a few short-term future decisions when planning is successful (§\hyperref[sec:61]{6.1}). For history step causality, we prevent the information flow from history steps and explore the impact of different history steps on the final decision (§\hyperref[sec:62]{6.2}).

In summary, our contributions are as follows:
\begin{itemize}
\item To the best of our knowledge, this work is the first to investigate the planning interpretability mechanisms in large language models. We demonstrate the \textbf{Look-Ahead Planning Decisions Existence Hypothesis}.
\item We reveal that the internal representations of LLMs encode a few short-term future decisions to some extent when planning is successful. These look-ahead decisions are enriched in the early layers, with accuracy decreasing as planning steps increase.
\item We prove that MHSA mainly extracts information from spans of the goal states and recent steps. The output of MHSA in the middle layers at the last token can directly decode the correct decisions partially in planning tasks.
\end{itemize}

\section{Experimental Setup}
\begin{figure}[t]
  \includegraphics[width=\columnwidth]{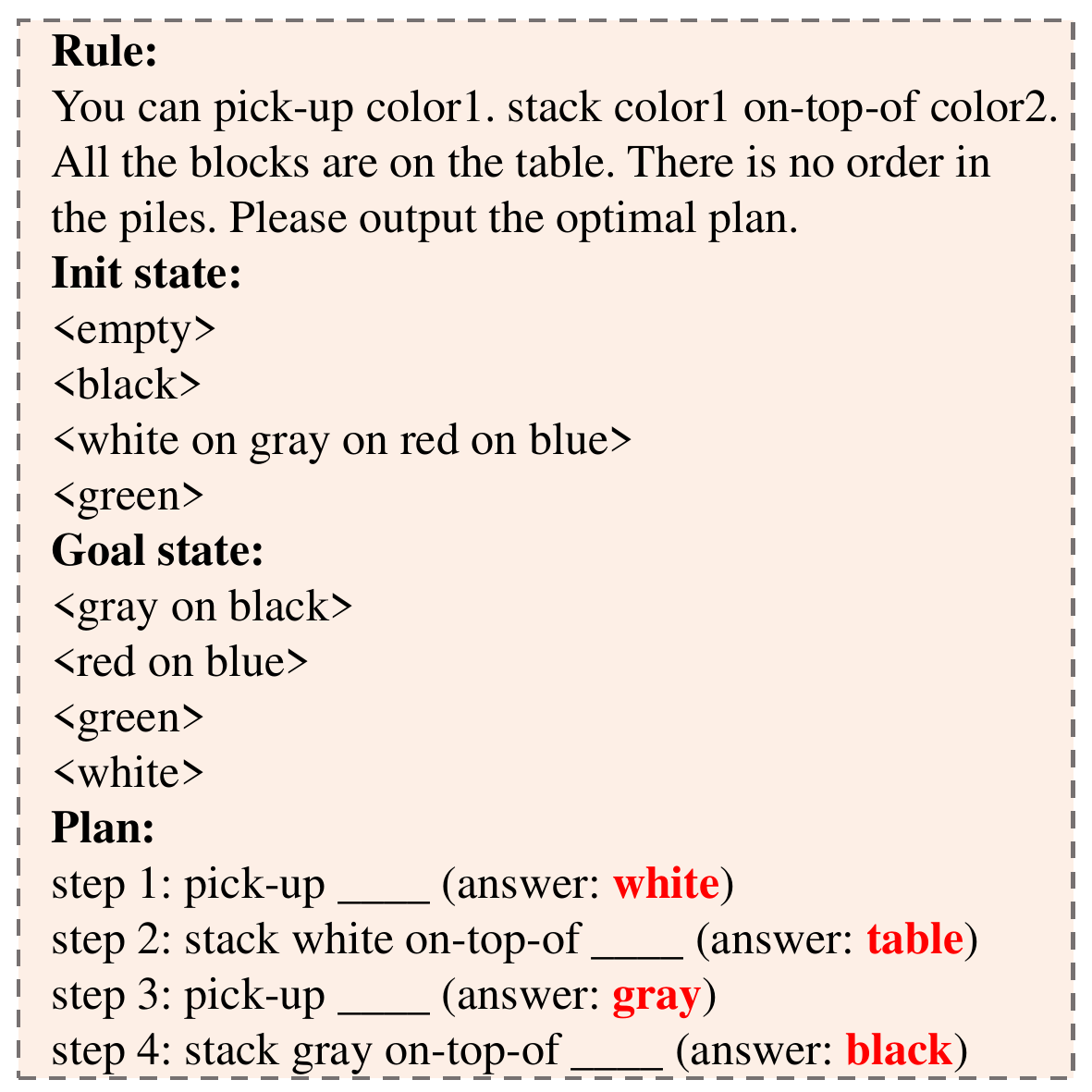}
  \caption{An example of Blocksworld.}
  \label{fig_task}
\end{figure}

In this paper, we study the Blocksworld task in a fully-observed setting where all entity states are known from the init state and goal state, so exploration is not needed \cite{zhang2024pddlego}. Given a rule $R$, an initial state $S_{\text{init}}$, a goal state $S_{\text{goal}}$, task description prompts $C$, the current step $t$, history $a_1 \ldots a_{t}$, model needs to predict the next action $a_{t+1}$ in accordance with its generative distribution $p(a_{t+1} \mid R, S_{\text{init}}, S_{\text{goal}}, C, a_1 \ldots a_{t})$ \cite{hao2023reasoning}. In this paper, all inputs are in text form. All inferences are performed using the teacher-forcing method. Previous evaluation works \cite{valmeekam2023planning} mainly involved generating a complete plan and then placing it into the environment for assessment. However, since our primary focus is on open-source models, we have reduced the difficulty by using a fill-in-the-blank format for evaluating the models.
An example is shown in Figure~\ref{fig_task}. The input mainly consists of four parts: rule, initial state, goal state, and plan.

\paragraph{Data} Previous Blocksworld evaluation benchmarks \cite{valmeekam2023planning} put the plans generated by models into an environment to verify the correctness. However, existing interpretability methods, such as information flow \cite{wang2023label}, require gold labels. Therefore, we synthesize a dataset containing optimal plans, with specific data statistics shown in Table~\ref{tab:data_table}. We generate data with 4, 5, and 6 color varieties, 4 piles, and a maximum of 6 steps, where pick-up and stack are considered as two different steps. There are three levels: LEVEL1 (L1) with two steps, LEVEL2 (L2) with four steps, and LEVEL3 (L3) with six steps. We choose the optimal path from the initial step to the final step. For samples with multiple optimal paths, we select one to include in the training set, ensuring that samples in the test set have unique optimal paths. We split the dataset into training and test sets with a ratio of 1:3.

\paragraph{Metric}  In the Blocksworld task, we use two metrics: single-step success rate and complete plan success rate. The single-step success rate evaluates whether each individual action is correct, defined as:
\begin{equation}
S_{\text{step}} = \frac{1}{N} \sum_{i=1}^{N} r_i  \\ 
\end{equation}
where $N$ is the total number of steps and $r_i$ indicates the success of the $i$-th step (1 for success, 0 otherwise). The complete plan success rate evaluates whether the entire planning process is correct, defined as:
\begin{equation}
S_{\text{plan}} = \frac{1}{M} \sum_{j=1}^{M} R_j\\ 
\end{equation} 
where $M$ is the total number of tested plans and $R_j$ indicates the success of the $j$-th plan (1 for complete success, 0 otherwise).

\paragraph{Model} We evaluate two large language models: Llama-2-7b-chat \cite{touvron2023llama} and Vicuna-7B \cite{vicuna2023}. Since open-source models have preliminary planning capabilities, we enhance the ability of large language models to complete planning tasks through training.

\paragraph{Experiment Setting} We conduct full parameter fine-tuning on Llama-2-7b-chat-hf and Vicuna-7B for 3 epochs. The training process involves a global batch size of 20, using the Adam optimizer with a learning rate of 5e-5. Llama-2-7b-chat-hf and Vicuna-7B achieve complete plan success rates of 61\% and 63\%, respectively, at LEVEL 3 with 6 blocks. We sample 400 correct data points from LEVEL 3 with 6 blocks for our analysis. We conduct experiment based on HuggingFace’s Transformers\footnote{\url{https://github.com/huggingface/transformers/}}, PyTorch\footnote{\url{https://github.com/pytorch/pytorch/}}, baukit \footnote{\url{https://github.com/davidbau/baukit/}} and pyvene\footnote{\url{https://github.com/stanfordnlp/pyvene}} \cite{wu2024pyvene}.

\section{Background}
A transformer-based language model begins by converting an input text into a sequence of $ N $ tokens, denoted as $ s_1, \ldots, s_N $. Each token $ s_i $ is mapped to a vector $ x_i^0 \in \mathbb{R}^d $. $ E \in \mathbb{R}^{|V| \times d} $ is the decoder matrix in the last layer, where $ V $ is the vocabulary, $d$ is embedding dimension. Each layer comprises a multi-head self-attention (MHSA) sublayer followed by a multi-layer perception (MLP) sublayer \cite{vaswani2017attention}. Formally, the representation $ x_i^{\ell} $ of token $ i $ at layer $ \ell$ can be obtained as follows:

\begin{equation}
\mathbf{x}_i^{\ell}=\mathbf{x}_i^{\ell-1}+\mathbf{attn}_i^{\ell}+\mathbf{m}_i^{\ell}
\end{equation}

$a_i^\ell$ and $m_i^\ell$ represent the outputs of the MHSA and MLP sub-layers of the $\ell$-th layer, respectively. By using $E$, an output probability distribution can be obtained from the final layer representation:
\begin{equation}
p_{i}^{L} = \text{softmax} \, (Ex_i^L)
\end{equation}

\section{Overview of Analysis}
We analyze the look-ahead planning mechanisms of the models from two stages. (1) In the first stage, we explore the internal mechanisms of this process in planning tasks from the perspectives of information flow and component functions. We demonstrate that the middle layer MHSA can directly decode the answers to a certain extent, and we prove that MHSA mainly extracts information from spans of the goal states and recent steps (§\hyperref[sec:5]{5}). (2) In the second stage, to determine the presence of future decisions, we employ the probing method to examine future decisions, uncovering that the intermediate and upper layers encode these decisions. Regarding the causality of historical steps, we inhibit the information flow from past steps and analyze the effects of different historical steps on the ultimate decision (§\hyperref[sec:6]{6}).

\section{Information Flow in Planning Tasks} \label{sec:5}
To trace the source of the correct answer, we begin with the last token. For example, in the first step "pick up", the last token is "up". The model should process the initial state, target state, and history of steps to decide which color to pick up, such as "blue". We analyze this process from two perspectives. (1) First, we study MLP and MHSA functions at the last token by extraction rates \cite{geva2023dissecting}. We find that the output of MHSA in the middle layers can directly decode the correct colors to a certain extent (§\hyperref[sec:51]{5.1}). (2) Based on this, we further trace the source of the correct colors by information flow \cite{wang2023label}. From the perspective of early and late planning stages, we prove that MHSA mainly extracts information from spans of the goal states and recent steps (§\hyperref[sec:52]{5.2}).

\begin{figure}[t]
  \includegraphics[width=\columnwidth, height=5cm]{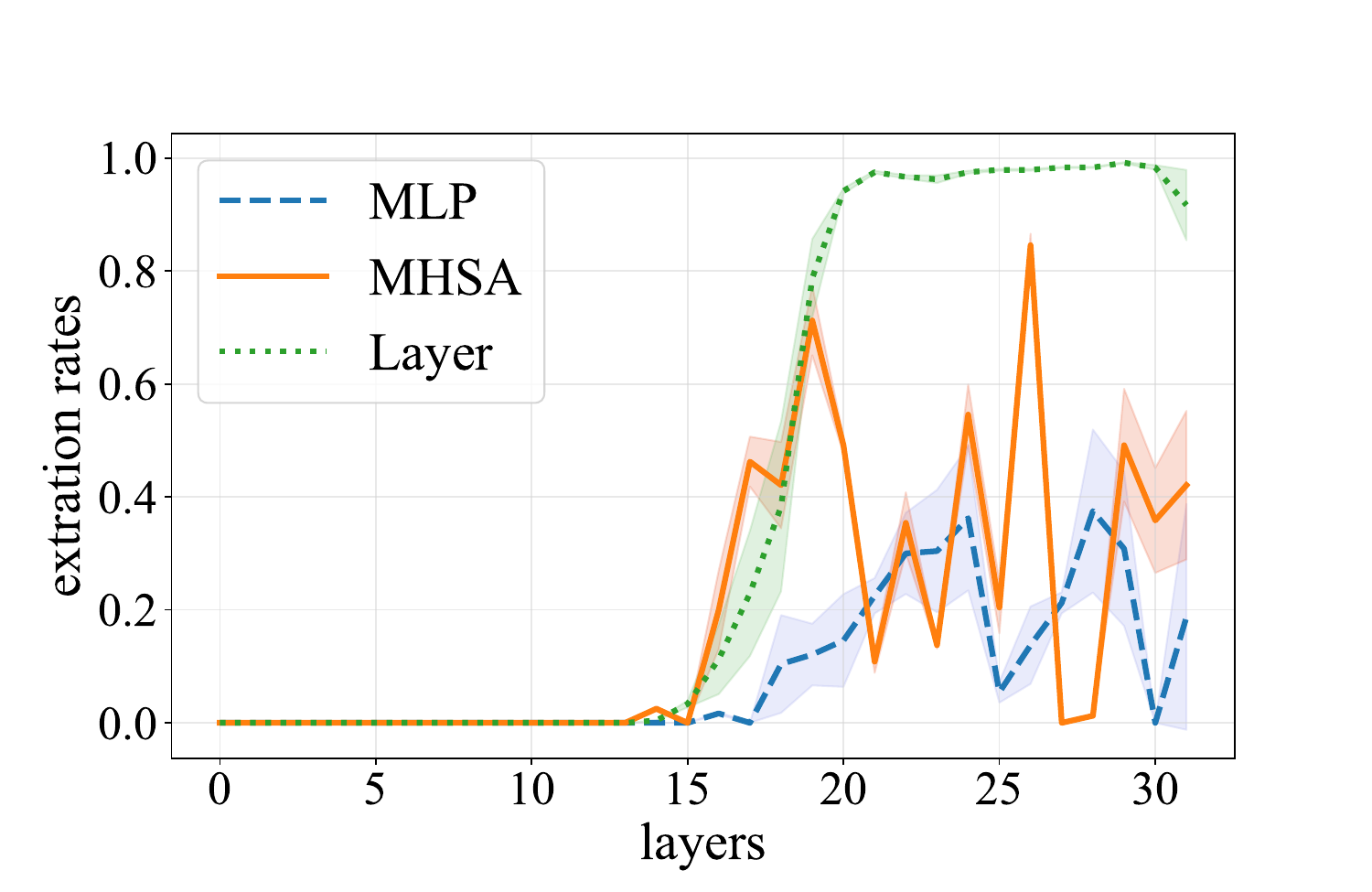}
  \caption{Extraction rate of different components in Llama-2-7b-chat-hf.} 
  \label{Figure-newex11_llama}
\end{figure}

\begin{figure}[t]
  \includegraphics[width=\columnwidth, height=5cm]{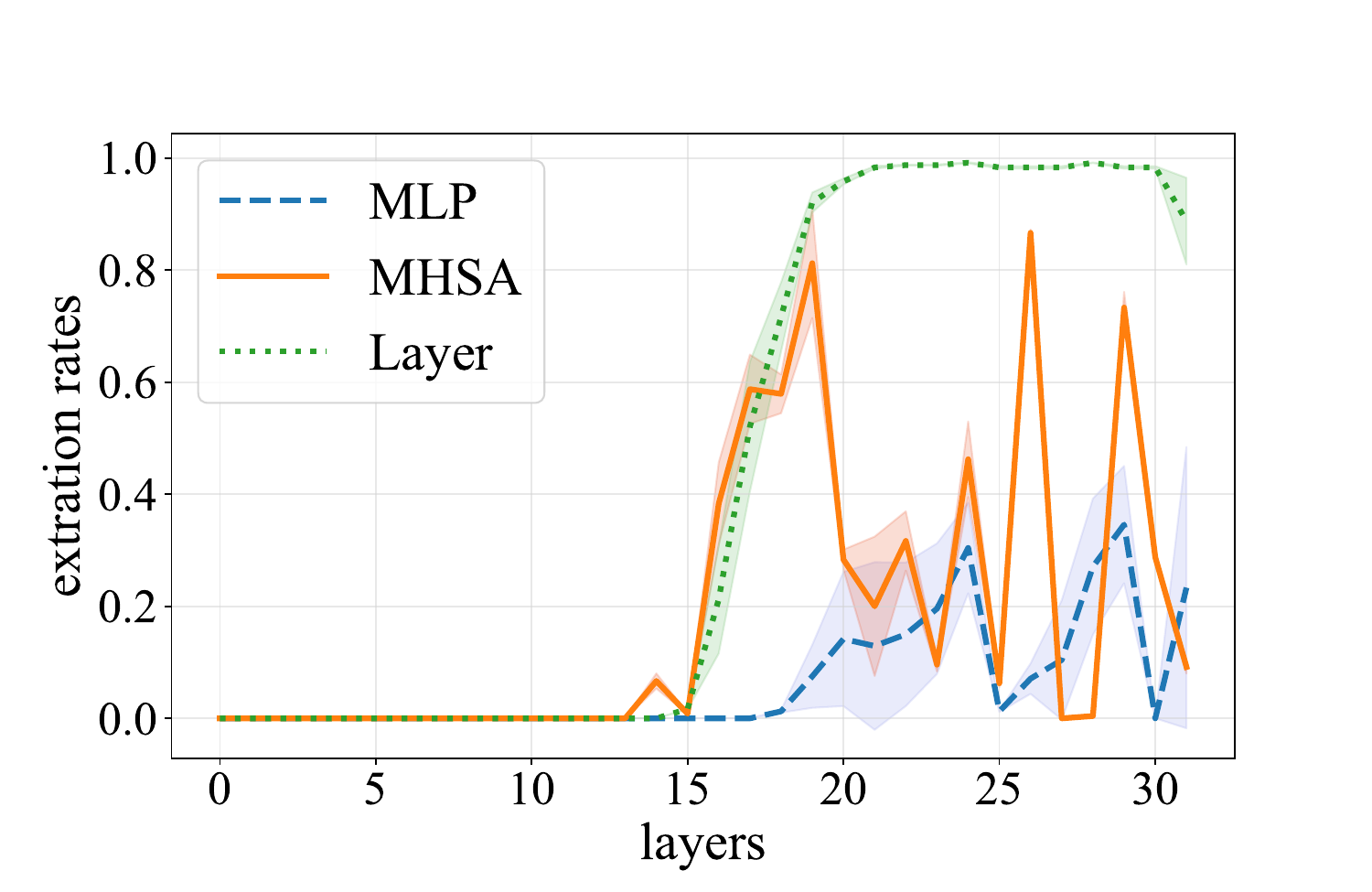}
  \caption{Extraction rate of different components in Vicuna-7b.} 
  \label{Figure-newex11_vicuna}
\end{figure}

\subsection{Attention Extract the Answers} \label{sec:51}
From the perspective of the model's internal components, we analyze the functions of different components of the models. The first question is how the model extracts answers from history. We start from the position of the last token and study the roles of the MLP and MHSA components in the answer generation process. Specifically, we investigate whether different components at different layers can directly decode the final answer.

\begin{figure*}[h]
  \includegraphics[width=\textwidth, height=8cm]{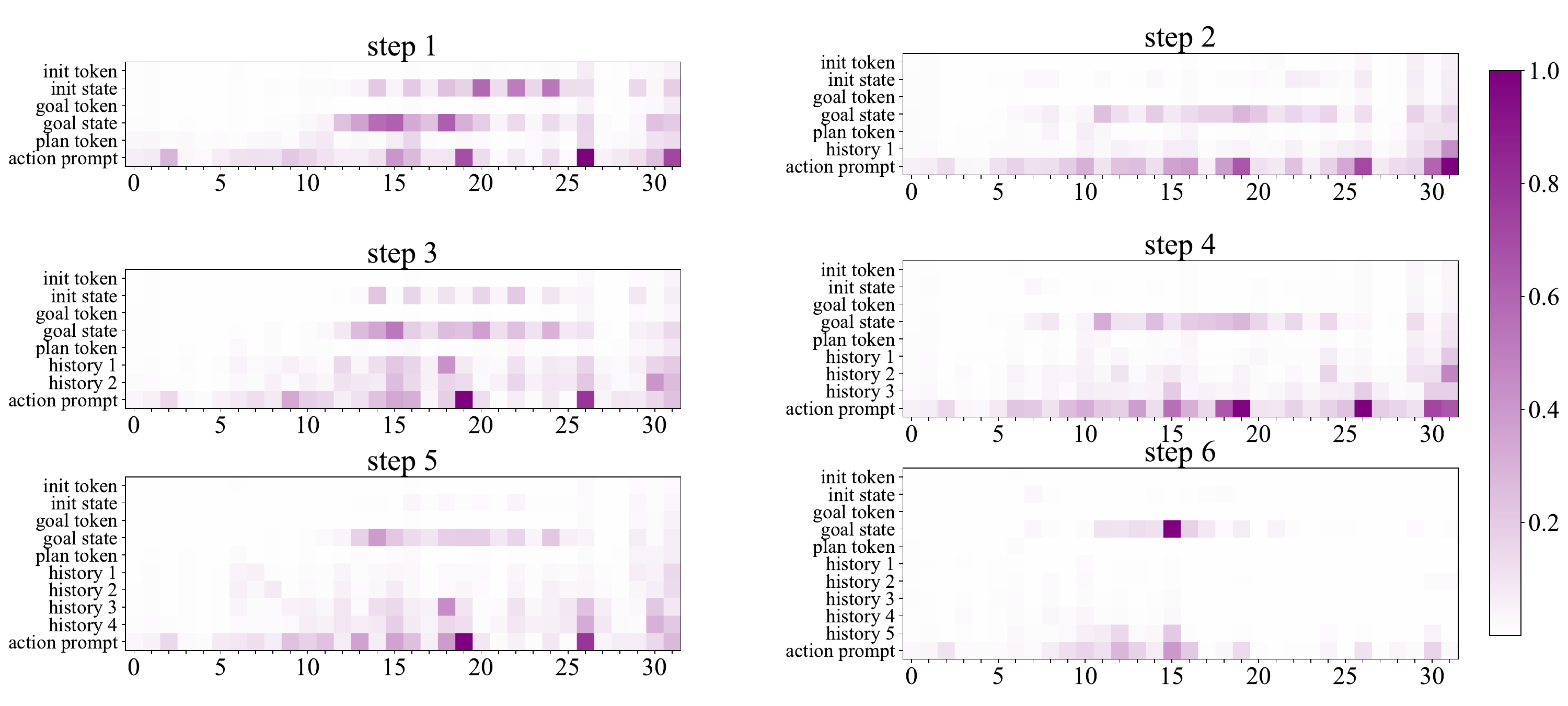}
  \caption{Information flow of last token in Llama-2-7b-chat-hf.}
  \label{Figure-newex12-llama}
\end{figure*}

\paragraph{Experiments} We use the extraction rate \cite{geva2023dissecting} to analyze the functions of different components. Specifically, we calculate the extraction rate:
\begin{equation}
    e^{\ast}:=arg max\left( p_{N}^{L}\right) \\ 
\end{equation}
\begin{equation}
    \widehat{e}:=argmax\left( Eh_{N}^{\ell}\right) \\
\end{equation}
In this equation, $h$ represents the internal representation of the MLP, MHSA and layer output, $N$ is the position of the last token,  $\ell$ is the layer of models, $\ell \in \left[ 1,L\right] $. When $ e^{\ast}$ = $\widehat{e}$, it is considered as an extraction event. We calculate the extraction rate of the last token for each layer for each step in the Blocksworld. We then compute the mean and variance of these rates.

\paragraph{Results and Analysis} As shown in Figure~\ref{Figure-newex11_llama} and Figure~\ref{Figure-newex11_vicuna}, we observe that (1) MHSA has a higher extraction rate compared to MLP, indicating that attention is primarily responsible for answer extraction. (2) Layer output gradually forms a stable answer in the middle to upper layers (from the 15th layer to the 20th layer). In these layers, the extraction rate of MHSA is significantly higher than MLP, suggesting that MHSA plays a major role during the decision-making period. (3) The variance in extraction rates across different steps is smaller for MHSA compared to MLP, indicating that MHSA layers show higher consistency across different steps.

\subsection{Attention Extract from Goal and History} \label{sec:52}
In the previous section, we discover that MHSA is responsible for extracting answers from the context, but which chunk to extract the answer from is still unclear. In this section, we decompose the input into several chunks to identify which chunk MHSA primarily relies on. We use the information flow method \cite{wang2023label}, first calculating the information flow at the token granularity, and then taking the average of different tokens within the same chunk to represent the information flow at the chunk granularity. This will help us locate the influence of different chunks on the last token.

\paragraph{Experiments} We calculate the information flow between layers. Specifically, for the input, we divide it into different chunks, including init token (which is "Init:"), init state (which is "<blue on red>"), target token, target state, six history steps (For step 1, which is "step 1: pick-up white"), action prompt (pick-up or stack on-top-of) and last token. We calculate the information flow $I_{token,\ell}$ for each token at the $\ell-th$ layer. The specific calculation method is as follows:
\begin{equation}
I_{token,\ell}=\left| \sum _{hd}A_{hd,l}\odot \frac{\partial L(x)}{\partial A_{h,\ell}}\right| 
\end{equation}

 Where $A_{hd,l}$ is the attention score of the $\ell$-th layer, $hd$ is the $hd$-th head, and $L(x)$ is the loss function. Here, we use $I(i, j)$ to represent the score flowing from token j to token i. Based on the token information flow, we calculate the chunk information flow, denoted as $I_{chunk,\ell}$: 
 \begin{equation}
I_{chunk,\ell} = \frac{\sum_{i=k_1}^{k_2} \sum_{j=t_1}^{t_2} I_{token,\ell}(i, j)}{(k_2-k_1+1)(t_2-t_1+1)} 
\end{equation}
 Specifically, we consider the information flow from the span $[k1, k2]$ of a chunk $k$ to the span $[t1, t2]$ of another chunk $t$. We calculate the average of information flow from chunk $k$ to $t$. Due to the causal attention, we only compute the information flow for the lower triangular matrix. We calculate the chunk information flow for each prediction step.

\paragraph{Results and Analysis} The results are shown in Figure~\ref{Figure-newex12-llama} and Figure~\ref{Figure-newex12-vicuna}. The vertical axis represents the information flow from the chunk to the last token. The horizontal axis represents the information flow at layer $\ell$. The values inside represent the scores of information flow. We calculate the information flow for six decision steps. It shows that: (1) In steps 1 to 6, the goal states are highlighted at each step. This indicates that MHSA extracts information from the goal state at each stage, demonstrating that it mainly relies on goal states. (2) Taking the step 5 as an example, history 3 and history 4 are more prominent compared to history 1 and history 2. It reveals that MHSA also mainly relies on recent history rather than earlier spans of steps.

\section{Internal Representations Encode Planning Information} \label{sec:6}
Based on the previous sections, we discover that MHSA directly extracts answers from the context, but it is still unclear what information is encoded in internal representations. In this section, we demonstrate the look-ahead capability of models from both future decisions existence and history step causality perspectives. (1) For future decisions existence, we use the probing method to probe each layer of the main positions in the context. We find that the accuracy of the current state information gradually decreases as the steps progress. We also find that the middle and upper layers encode future decisions with accuracy decreasing as planning steps increase, proving the look-ahead planning hypothesis (§\hyperref[sec:61]{6.1}). (2) For history step causality, we employ a method that involves setting certain information keys of MHSA to zero. We find there is still a probability of generating the correct answer by relying solely on a single step, but it's difficult to support plan for the long-term (§\hyperref[sec:62]{6.2}).

\subsection{Internal Representations Encode Block States and Future Decisions}  \label{sec:61}
In this section, we analyze what information is encoded in the internal representations within the information flow and how this information evolves layer by layer. We examine whether the internal representations encode two types of information \cite{li2022emergent,pal2023future}: \textit{Current Block States} and \textit{Future Decisions}. 
\textit{Current Block States} refer to the state of the blocks at step $t$. For example, in Figure~\ref{fig_introduction}, when following the green path, the \textit{Current Block State} initially starts in the $S_{\text{init}}$. After executing the first and second steps, the internal representation of the \textit{Current Block State} transitions from the $S_{\text{init}}$ to $S_{12}$.
\textit{Future Decisions} refer to the information about future decisions at step $t$. For example, in Figure~\ref{fig_introduction}, when following the green path and executing the first step (blue), the question is whether the model's internal representation already stores information about future decisions (red, yellow, blue).

\begin{figure}[h]
  \includegraphics[width=\columnwidth]{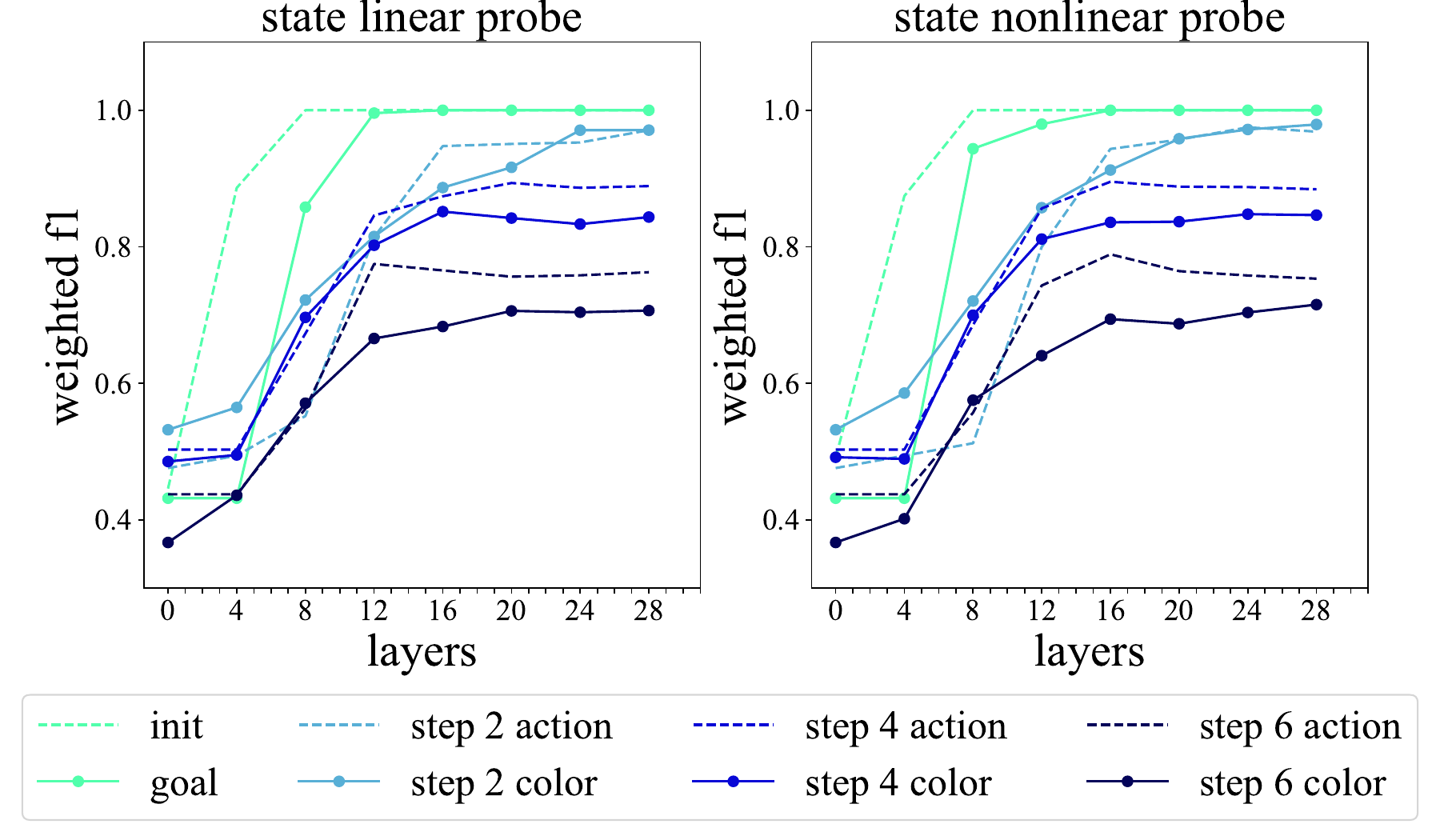}
  \caption{State probe in Llama-2-7b-chat-hf.}
  \label{Figure-newex212-llama}
\end{figure}

\begin{figure}[h]
  \includegraphics[width=\columnwidth]{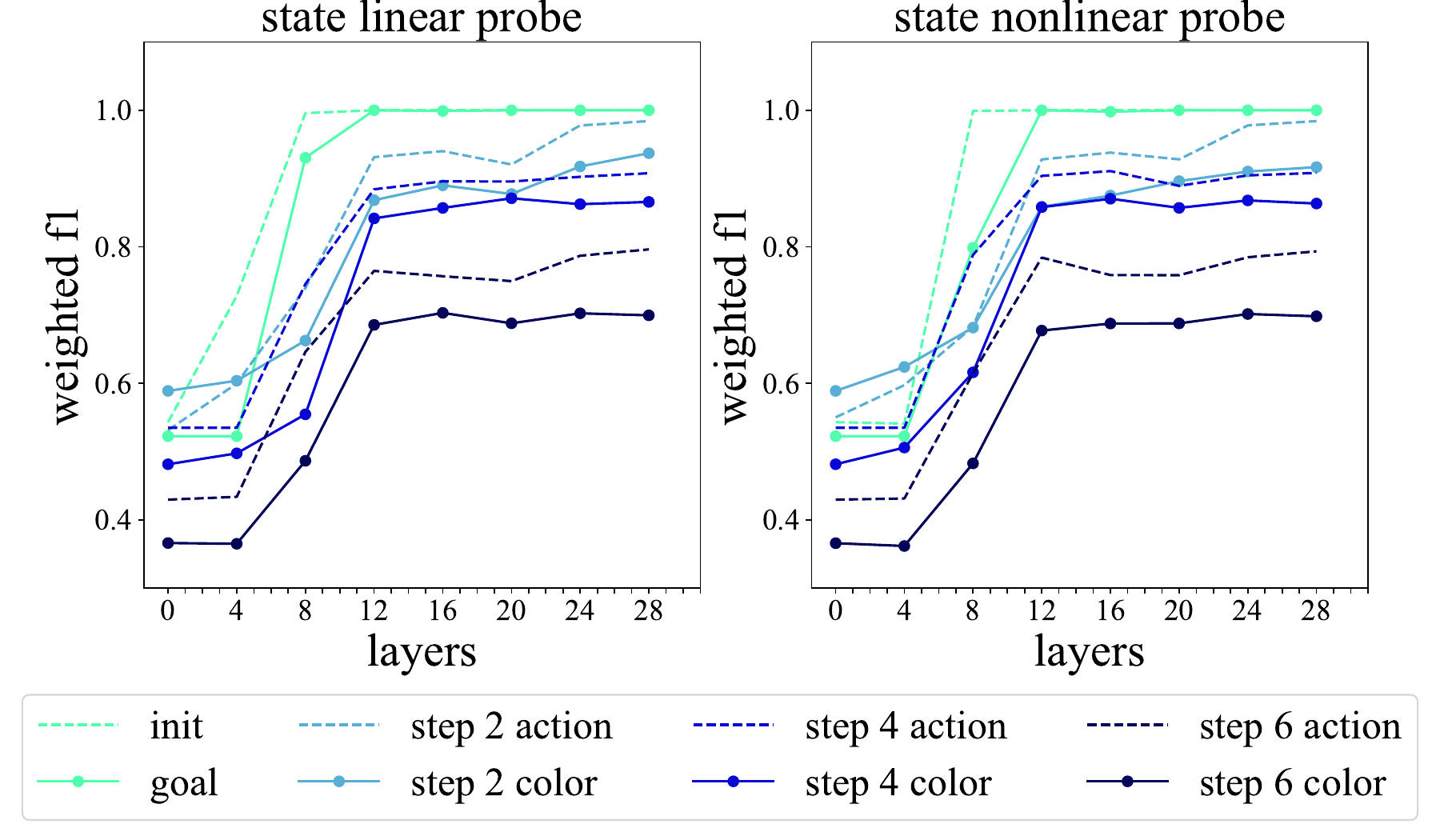}
  \caption{State probe in Vicuna-7b.}
  \label{Figure-newex212-vicuna}
\end{figure}

\paragraph{Experiments} We probe internal representations of the initial state, goal state, and steps with layer $\ell \in \left[ 1,L\right] $. We train linear probes and nonlinear probes for each chunk and each layer. A linear probe can be represented as $p_{\theta}(x_n^\ell) = \text{softmax}(W x_n^\ell)$. And a nonlinear probe can be described as $p_{\theta}(x_n^\ell) = \text{softmax}(W_1 \, \text{ReLU}(W_2 x_n^\ell))$. Using the linear probe as an example, we consider six steps and six colors of blocks. For \textit{Current Block States}, the input to the probe is a hidden layer representation $h$ of the model. The output is a 12x8 matrix representing probabilities, where 12 denotes the colors of the blocks above and below each color block, and 8 represents 6 colors plus sky and table. For \textit{Future Decisions}, the input to the probe is $h$. The output is six predicted colors from steps 1 to 6, we only consider future steps in our evaluation. We split the training and test sets in a 4:1 ratio for 400 samples. For the evaluation, we calculate the weighted F1 accuracy for \textit{Current Block States} and accuracy for \textit{Future Decisions}.

\begin{figure}[h]
  \includegraphics[width=\columnwidth]{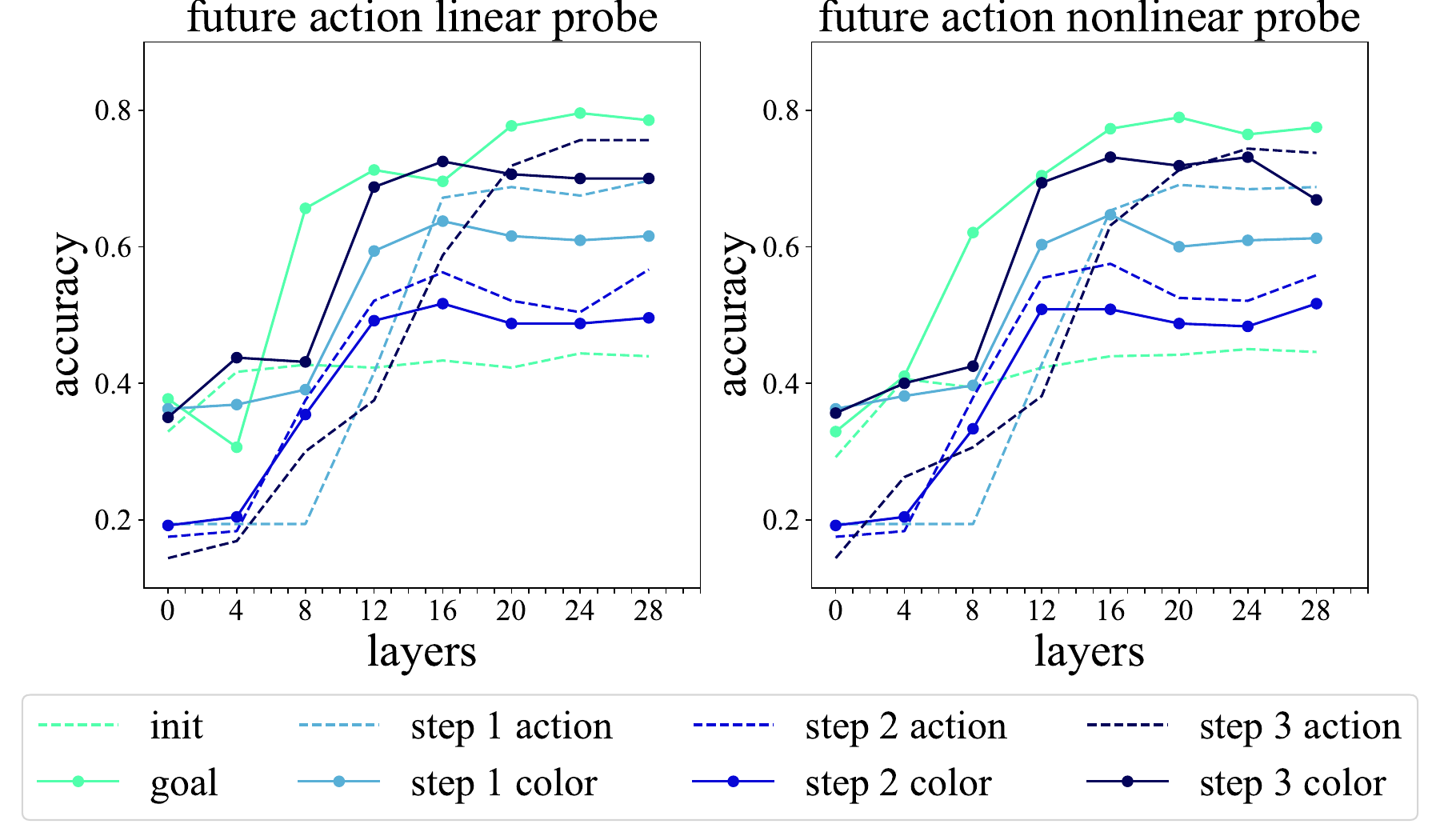}
  \caption{Action probe in Llama-2-7b-chat-hf.}
  \label{Figure-newex212-action-llama}
\end{figure}

\begin{figure}[h]
  \includegraphics[width=\columnwidth]{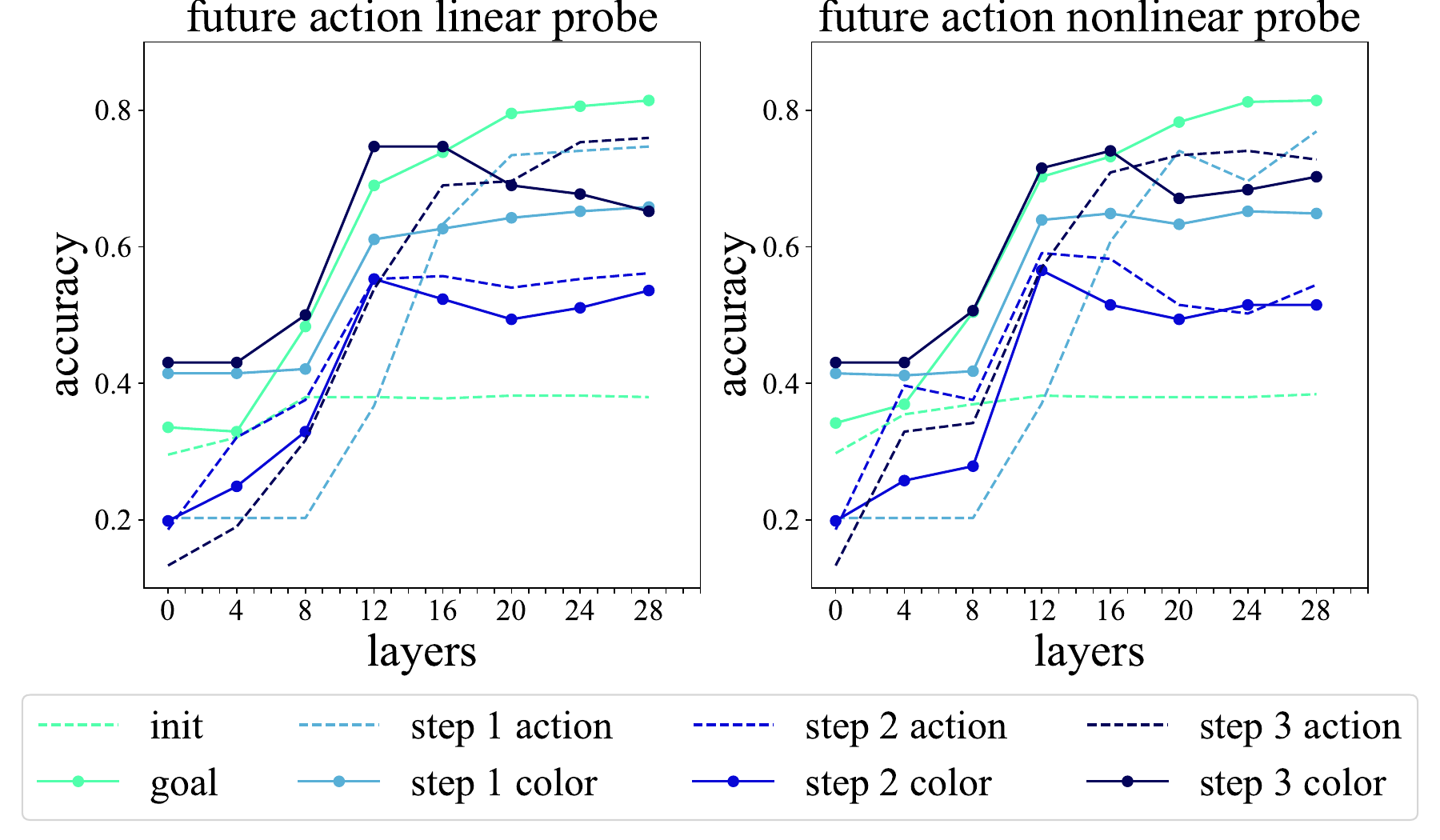}
  \caption{Action probe in Vicuna-7b.}
  \label{Figure-newex212-action-vicuna}
\end{figure}

\paragraph{Results and Analysis} As shown in Figure~\ref{Figure-newex212-llama} and Figure~\ref{Figure-newex212-vicuna}, the horizontal axis represents the layers probed, while the vertical axis represents the mean accuracy of the probe test. Different colored lines represent the probed spans of states and steps. (1) We observe that as the number of layers increases, the accuracy of the probe gradually improves. This indicates that the early layers of the model are enriching the representation of the current state. (2) The black line (step 6) in the figure has a lower accuracy compared to the light blue line (step 2), demonstrating that as the planning steps progress, the models are difficult to maintain the representations of the current placement of the blocks. (3) By comparing the linear probe in Figure~\ref{Figure-newex212-llama} and the nonlinear probe in Figure~\ref{Figure-newex212-vicuna}, we find that both have the same trend, indicating that the model internally stores the current state in a linear manner. A similar trend in \textit{Future Decisions} is shown in Figure~\ref{Figure-newex212-action-llama} and Figure~\ref{Figure-newex212-action-vicuna} for actions. It reveals that look-ahead decisions are enriched in the early layers.

\begin{figure}[h]
    \centering
  \includegraphics[width=0.8\columnwidth]{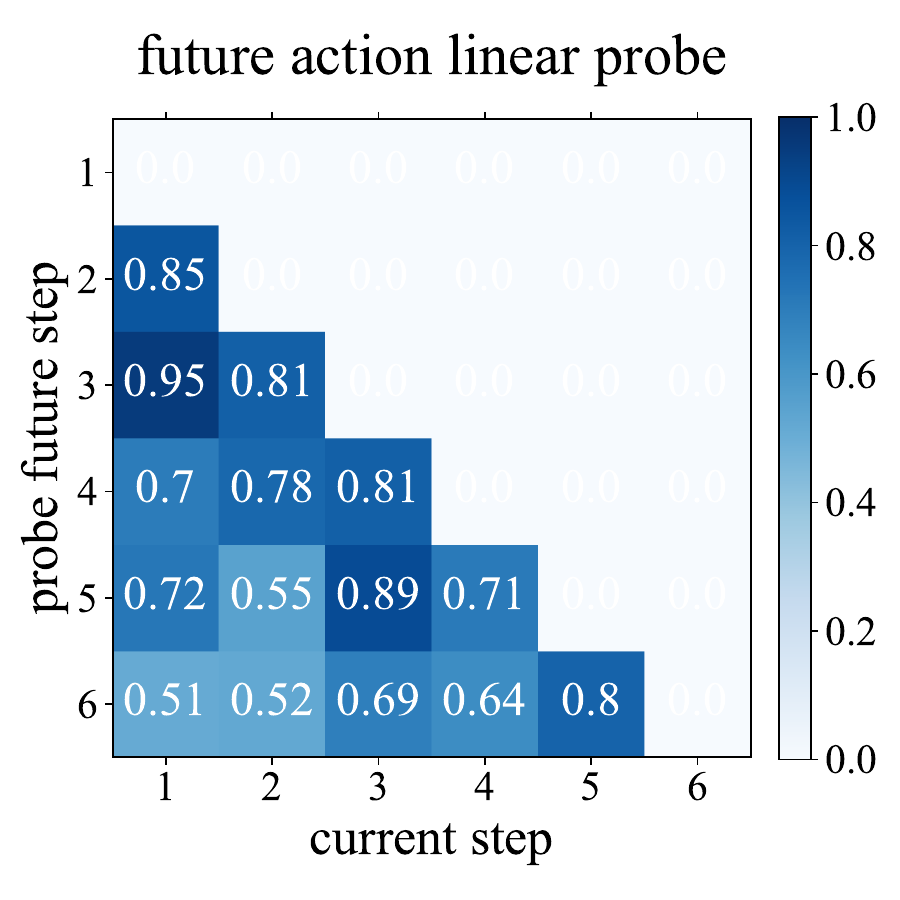}
  \caption{Future action linear probe in Llama-2-7b-chat-hf.}
  \label{Figure-newex214-llama}
\end{figure}

\paragraph{Supplementary Analysis} As shown in Figure~\ref{Figure-newex214-llama} and Figure~\ref{Figure-newex214-llama—non}. They illustrate the accuracy of future decisions based on the current step. Each column represents the current step, while the rows represent the max accuracy of the probe in predicting future answers. We observe the following : (1) For the sixth row and first column, the probe can predict the future sixth step with an accuracy of 0.51 at the first step. This indicates that the model stores information about future decisions in advance, supporting the hypothesis of forward planning. (2) For each row, the values increase from left to right. For example, the accuracy in the fifth column of the sixth row is higher than that in the first column. This means the model is more certain about the output of the sixth step at the fifth step compared to the first step, demonstrating that the model has difficulty in planning over long distances. (3) The accuracy for the first column, representing the prediction accuracy for the next five steps after the initial step, shows a declining trend, indicating that the model stores future decision information in advance, supporting the hypothesis of look-ahead planning decisions existence hypothesis.

\subsection{Internal Representations Facilitate Future Decision-making} \label{sec:62}
In this section, we further verify the causal effect of planning information at different steps. We test the causality between planning information in the previous history \( t_a \) and decisions in step \( t_b \), where \( t_a < t_b \). Specifically, we compare whether the information from step \( t_1 \) contributes to the planning in step \( t_2 \). If the model is greedy in its planning, there should be no decision information in \( t_a \) that can help make better decisions in \( t_b \). Therefore, we set the key of MHSA in historical decision \( t \) to 0 to study the causal effect of historical information on future predictions.

\paragraph{Experiments} For each step \( t \), we have a history \( H_t = [a_1, a_2, ..., a_{t-1}] \), where each step span $a_i$ contains color tokens.

(1) Mask all steps: First, identify all color tokens in \( H_t \), and set the keys to 0 for these colors in each layer of MHSA, resulting in the masked historical information \( H'_t \). The main goal is to stop past decision information from affecting the current decision of the last token. Obtain the decision probability  \( y'_t \) based on \( H'_t \) in $t$ step.

(2) Make one step visible: Based on \( H'_t \), make only the color at position \( i \) visible, while masking the other positions, resulting in \( H''_{t,i} \). Use \( H''_{t,i} \) for prediction, Obtain the decision probability \( y''_{t,i} \).

(3) Calculate one step effect: Compare the decision probabilities obtained from masking all steps and from making one step visible to calculate the effect of a single step. The larger this value, the greater the impact of step \(i\) on step \(t\):
\begin{equation}
\text{Impact}_{i,t} = y''_{t,i} - y'_t
\end{equation}

\begin{figure}[t]
    \centering
  \includegraphics[width=0.8\columnwidth]{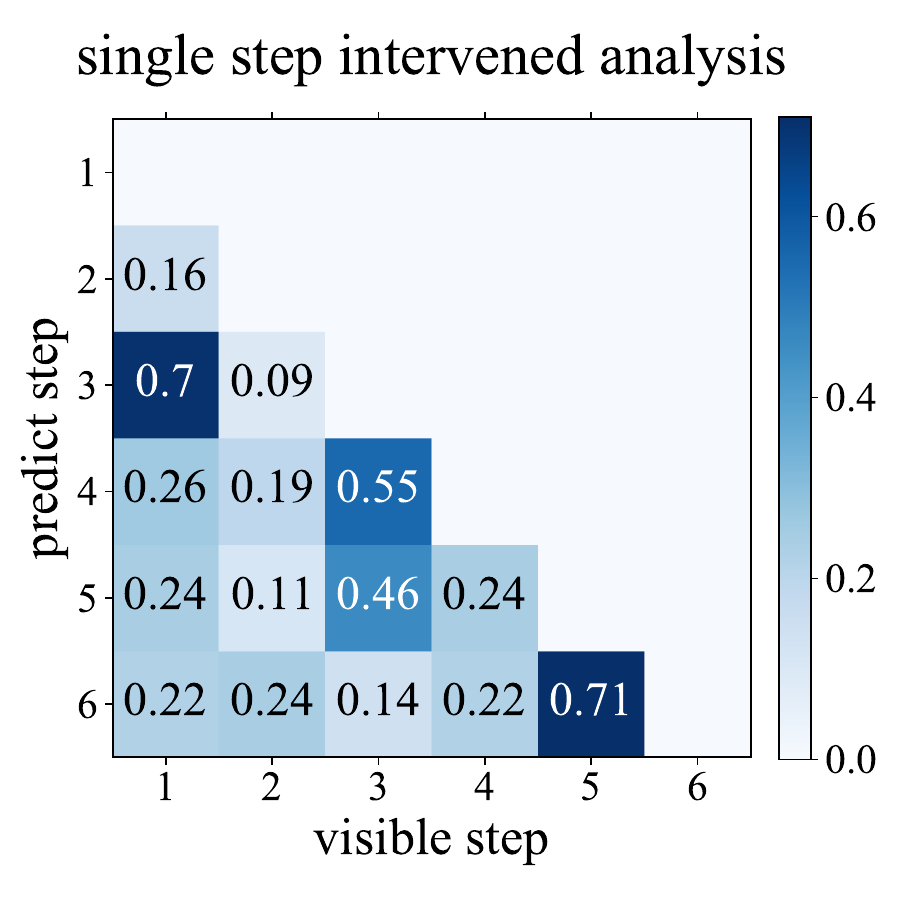}
  \caption{Single step intervened analysis in Vicuna-7b.} 
  \label{Figure-newex22-vicuna}
\end{figure}

\paragraph{Results and Analysis} As illustrated in the Figure ~\ref{Figure-newex22-vicuna} and Figure~\ref{Figure-newex22}, the columns represent the steps visible during prediction, the rows represent the steps being predicted, and the values inside represent the contribution of step $t$ to step $i$. (1) For example, in the second column of the sixth row, the model can increase the probability of inferring the correct decision in the sixth step by 0.24 just by using the information from the second step. This indicates that the model is not greedy and is not limited to only preparing for the next step, which causally proves the conclusion of look-ahead planning. (2) Observing the values in each column, for instance, the maximum value in the fifth row is 0.46, located in the third column. This represents that the third step is the most important for predicting the fifth step. It is found that the most important steps for prediction tend to be later steps, indicating that the look-ahead planning ability of LLMs is still relatively preliminary.

\section{Related work}
\paragraph{LLM-Based Agents} With the emergence of large language models, researchers begin to use them as intelligent agents \cite{xi2023rise,wang2024survey}. Significantly, ReAct \cite{yao2022react} innovatively combines CoT reasoning with agent actions. Some tasks utilize the planning capabilities of large language models through prompt engineering methods \cite{huang2022language,hao2023reasoning,yao2024tree,zhang2024pddlego}. Other researchers enhance the planning capabilities of large language models through fine-tuning methods \cite{zeng2023agenttuning,2024arXiv240605673Y}. Some researchers construct benchmarks to evaluate the planning ability of large language models \cite{shridhar2020alfworld,wang2022scienceworld,zhou2023webarena,deng2024mind2web,xu2023tool,qin2023toolllm}.

\paragraph{Mechanistic Interpretability} Recent works study mechanistic interpretability in factual associations, in-context Learning, and arithmetic reasoning tasks from the perspective of information flow \cite{geva2023dissecting,wang2023label,stolfo2023mechanistic}. Researchers also study Othello \cite{li2022emergent,nanda2023emergent}, chess \cite{karvonen2024emergent} and Blocksword \cite{wang2024alpine} in transformer. However, research on the mechanistic interpretation of large language models performing planning tasks is still unexplored. Our work conducts a preliminary study from the perspective of information flow and internal representation.

\paragraph{Look-Ahead} \citet{pal2023future,wu2024language,jenner2024evidence} demonstrate that it is possible to decode future tokens from the hidden representations of a language model at previous token positions. In task planning, a model needs to have look-ahead capabilities. However, it is not yet clear whether LLMs use similar mechanisms when planning. Our work focuses on the look-ahead mechanisms in planning in LLMs.

\section{Conclusion}
In this paper, we investigate the mechanisms of look-ahead planning in LLMs through the perspectives of information flow and internal representations. We demonstrate \textbf{Look-Ahead Planning Decisions Existence Hypothesis}. Our findings indicate that internal representations of LLMs encode a few short-term future decisions to
some extent when planning is successful. These look-ahead decisions are enriched in the early layers, with their accuracy diminishing as the number of planning steps increases. We demonstrate that MHSA mainly extracts information from the spans of goal states and recent steps. Additionally, the output of MHSA in the middle layers at the final token can partially decode the correct decisions.

\section*{Limitation}
Although our work provides an in-depth analysis and explanation of look-ahead planning mechanisms of large language models, there are several limitations. First, our analytical methods require access to the internal parameters and representations of open-source models. Although black-box large language models such as ChatGPT possess strong planning capabilities, we cannot access their internal parameters, making it challenging to interpret the most advanced language models. Second, our research primarily focuses on the planning mechanisms in Blocksworld. However, many other planning tasks, such as commonsense planning (e.g., "how to make a meal"), lack standard answers, making it difficult to evaluate the correctness of the planning and conduct quantitative analysis. We leave these limitations for future work.

\bibliography{arxiv}

\appendix
\label{sec:appendix}

\section{Additional Results}

\begin{figure*}[h]
  \includegraphics[width=\textwidth, height=8cm]{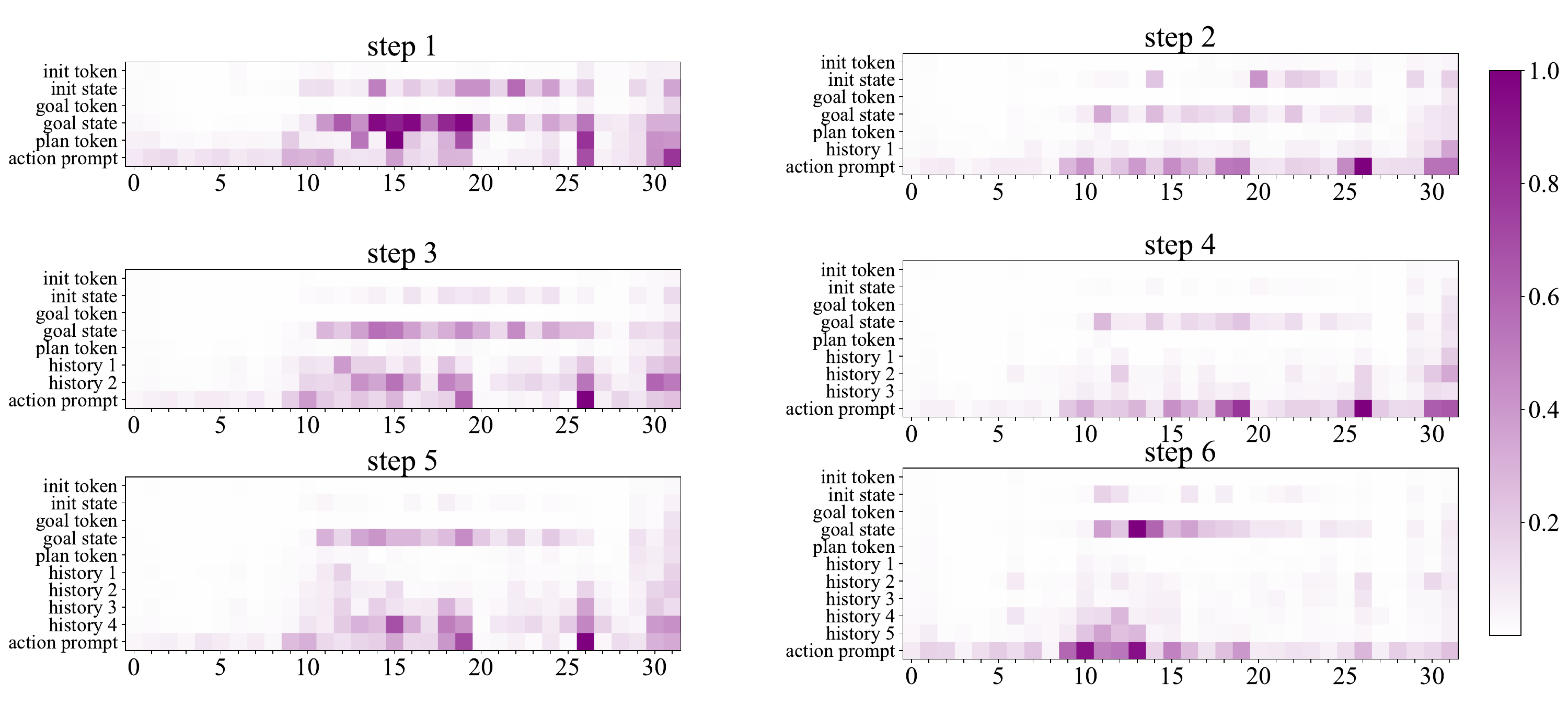}
  \caption{Information flow of last token in Vicuna-7b.}
  \label{Figure-newex12-vicuna}
\end{figure*}

\begin{figure}[h]
    \centering
  \includegraphics[width=0.8\columnwidth]{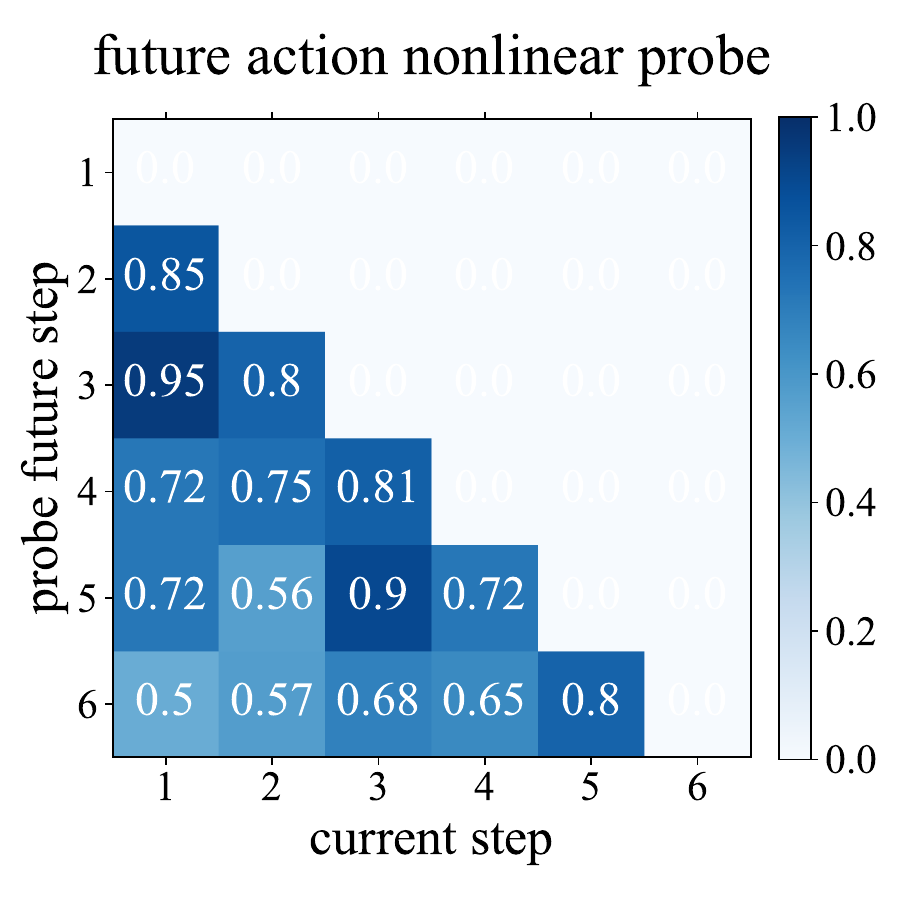}
  \caption{Future action nonlinear probe in Llama-2-7b-chat-hf.}
  \label{Figure-newex214-llama—non}
\end{figure}

\begin{figure}[h]
    \centering
  \includegraphics[width=0.8\columnwidth]{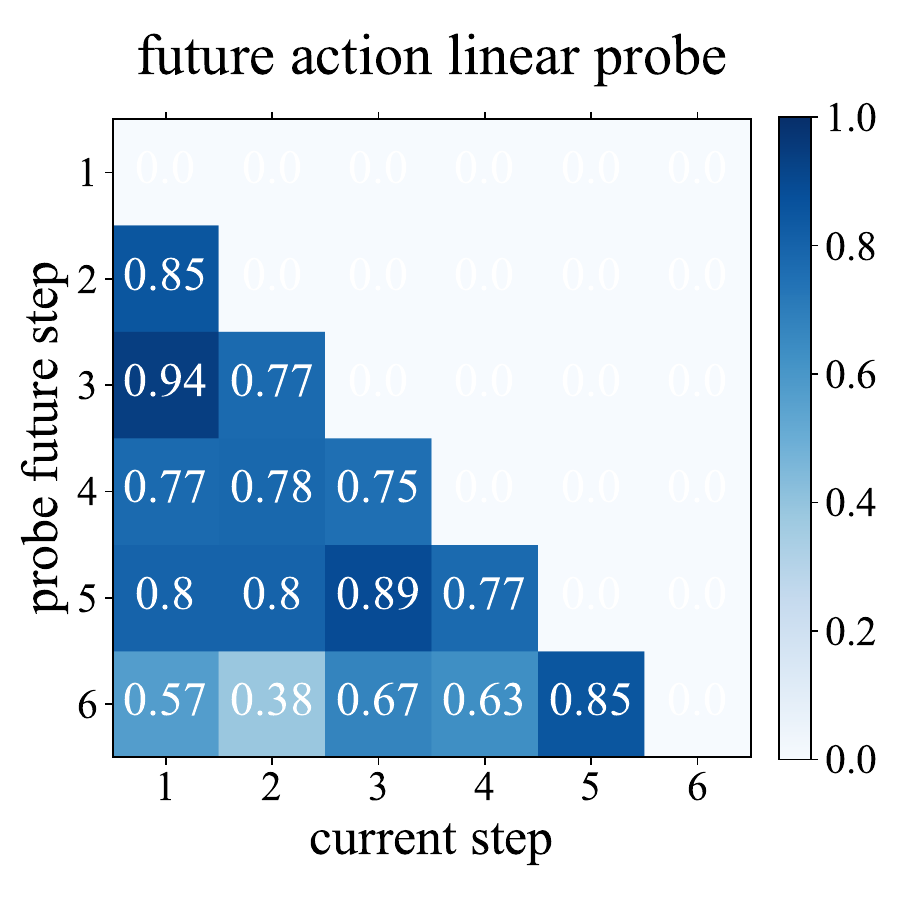}
  \caption{Future action linear probe in Vicuna-7b}
  \label{Figure-newex214-vicuna}
\end{figure}

\begin{figure}[h]
    \centering
  \includegraphics[width=0.8\columnwidth]{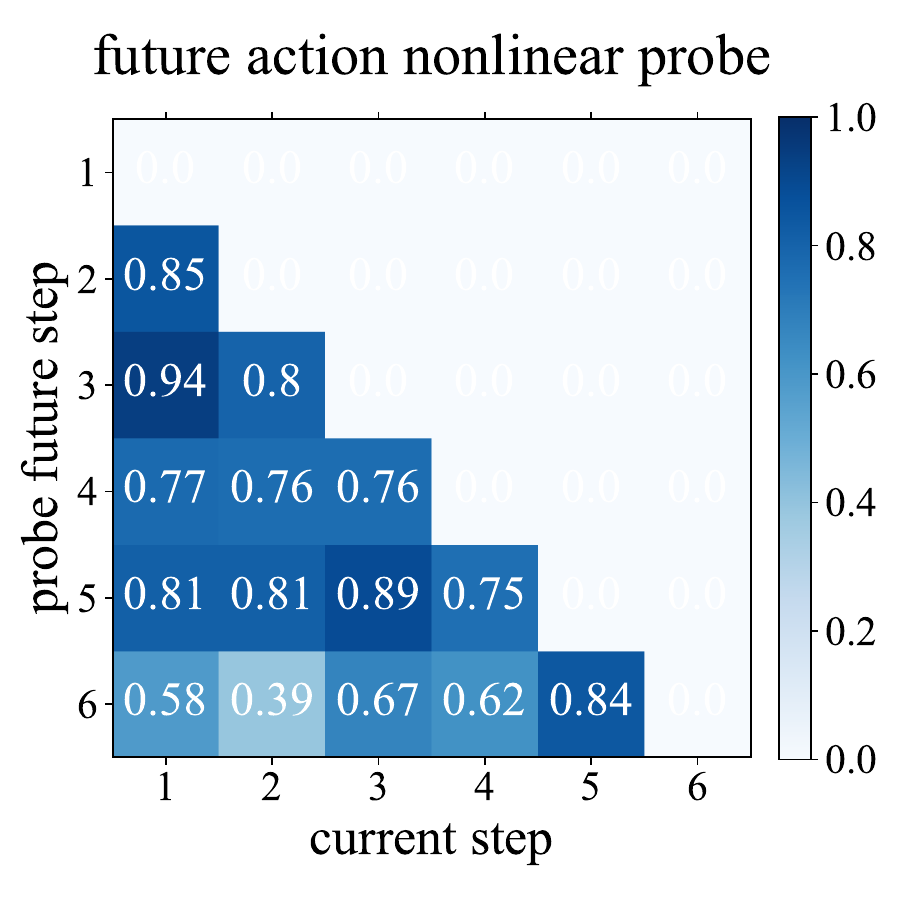}
  \caption{Future action nonlinear probe in Vicuna-7b}
  \label{Figure-newex214-vicuna—non}
\end{figure}

\begin{figure}[t]
    \centering
  \includegraphics[width=0.8\columnwidth]{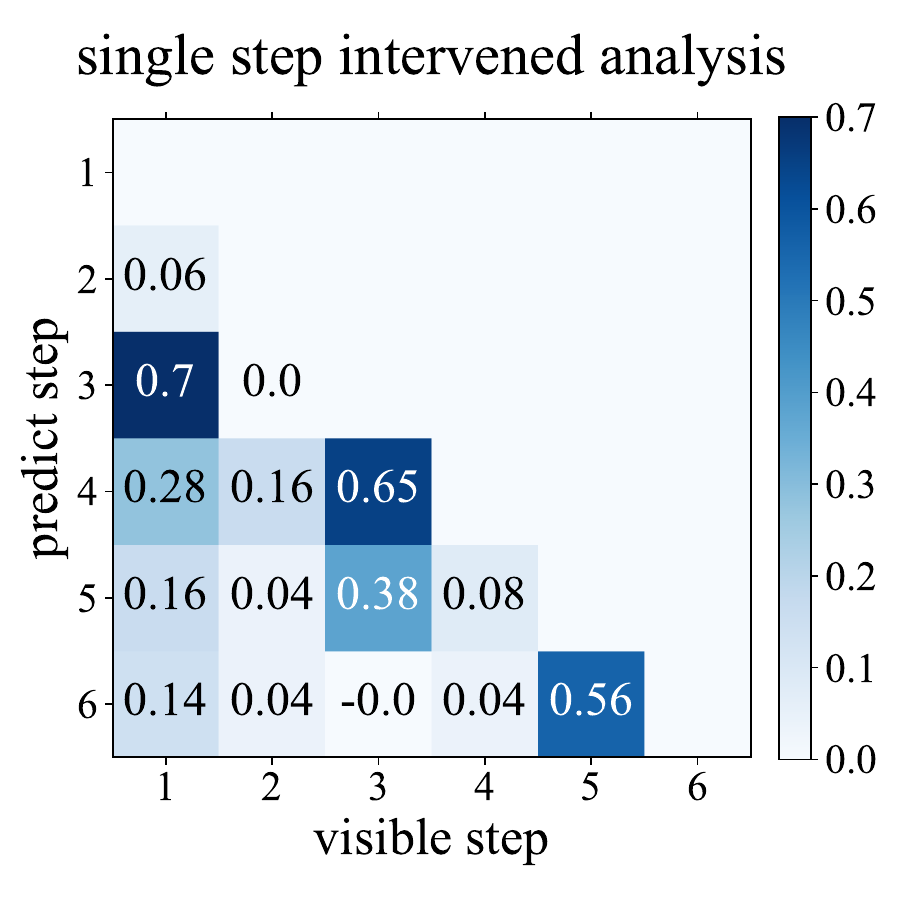}
  \caption{Single step intervened analysis in Llama-2-7b-chat-hf.} 
  \label{Figure-newex22}
\end{figure}

Information flow of last token in Vicuna-7b is shown in Figure~\ref{Figure-newex12-vicuna}. Future action nonlinear probe in Llama-2-7b-chat-hf is shown in Figure~\ref{Figure-newex214-llama—non}. Future action linear probe in Vicuna-7b is shown in Figure~\ref{Figure-newex214-vicuna}. Future action nonlinear probe in Vicuna-7b is shown in Figure~\ref{Figure-newex214-vicuna—non}, Single step intervened analysis in Llama-2-7b-chat-hf is shown in Figure~\ref{Figure-newex22}.

\begin{table}[h!]
  \centering
  \begin{tabular}{c|cccc}
    \hline
    \textbf{LEVEL}& \textbf{L1} & \textbf{L2} & \textbf{L3} & \textbf{Total} \\
    \hline
    \multicolumn{5}{c}{\textit{Train Size}} \\ 
        \hline
    4 blocks & 3 & 17 & 25 & 45 \\
    5 blocks & 1 & 23 & 121 & 145 \\
    6 blocks & 3 & 48 & 326 & 377 \\
    Total & 7 & 88 & 472 & 567 \\
    \hline
    \multicolumn{5}{c}{\textit{Test Size}} \\
        \hline
    4 blocks & 24 & 60 & 80 & 164 \\
    5 blocks & 34 & 115 & 268 & 417 \\
    6 blocks & 57 & 232 & 709 & 998 \\
    Total & 115 & 407 & 1057 & 1579 \\
    \hline
  \end{tabular}
  \caption{Blocksworld dataset statistics}
  \label{tab:data_table}
\end{table}

\end{document}